\documentclass[sigconf]{acmart}

\AtBeginDocument{%
  \providecommand\BibTeX{{%
    \normalfont B\kern-0.5em{\scshape i\kern-0.25em b}\kern-0.8em\TeX}}}

\renewcommand\footnotetextcopyrightpermission[1]{} 
\begin{document}

\title{Constructing Information-Lossless Biological Knowledge Graphs from Conditional Statements}

\author{Tianwen Jiang$^{1,2}$, Tong Zhao$^1$, Bing Qin$^2$, Ting Liu$^2$, Nitesh V. Chawla$^1$, Meng Jiang$^1$}

\affiliation{
    \institution{$^1$Department of Computer Science and Engineering, University of Notre Dame, Notre Dame, IN 46556, USA}
	\institution{$^2$Research Center for Social Computing and Information Retrieval, Harbin Institute of Technology, China}
}
\email{twjiang@ir.hit.edu.cn, tzhao2@nd.edu, {bqin, tliu}@ir.hit.edu.cn, {nchawla, mjiang2}@nd.edu}

\renewcommand{\shortauthors}{T. Jiang, T. Zhao, B. Qin, T. Liu, N. Chawla, M. Jiang}

\begin{abstract}
Conditions are essential in the statements of biological literature. Without the conditions (e.g., environment, equipment) that were precisely specified, the facts (e.g., observations) in the statements may no longer be valid. One biological statement has one or multiple fact(s) and/or condition(s). Their subject and object can be either a concept or a concept's attribute. Existing information extraction methods do not consider the role of condition in the biological statement nor the role of attribute in the subject/object. In this work, we design a new tag schema and propose a deep sequence tagging framework to structure conditional statement into fact and condition tuples from biological text. Experiments demonstrate that our method yields a information-lossless structure of the literature.
\end{abstract}




\maketitle

\section{Introduction}
Extracting information from biological text plays an important role in biological knowledge graph construction, relational inference, hypothesis generation and validation. Open IE systems extract a diverse set of relational tuples without requiring any relation-specific schema in advance, which was supposed to be ideally suited to the biological corpora. For example, given a statement sentence in a biochemistry paper~\cite{tomilin2016trpv5}: ``\textit{We observed that ... alkaline pH increases the activity of TRPV5/V6 channels in Jurkat T cells.}'', an Open IE system~\cite{stanovsky2018supervised} would return a (subject, relation, object)-tuple (\textit{alkaline pH, increases, activity of TRPV5/V6 channels in Jurkat T cells}).

Such an information representation has two problems. First, the condition that the channels were in the Jurkat T cells, on which the observation was obtained, remained unstructured in the object argument. In biological domains, conditions are essential in claiming observations and hypotheses~\cite{miller1947nature}. Second, the object name was long, infrequent, and less likely to be associated with other tuple extractions, leading to knowledge graph sparsity and poor inference, though the concept ``TRPV5/V6 channels'' and its attribute ``activity'' could be frequently mentioned. Unlike the general domain where relations are often linking two named entities, in biological domains, the relations could be linking either biological concepts or a concept's particular attributes. A well-designed information structure of biological statement is desired.

\begin{figure}[t]
    \centering
    {\includegraphics[width=0.48\textwidth]{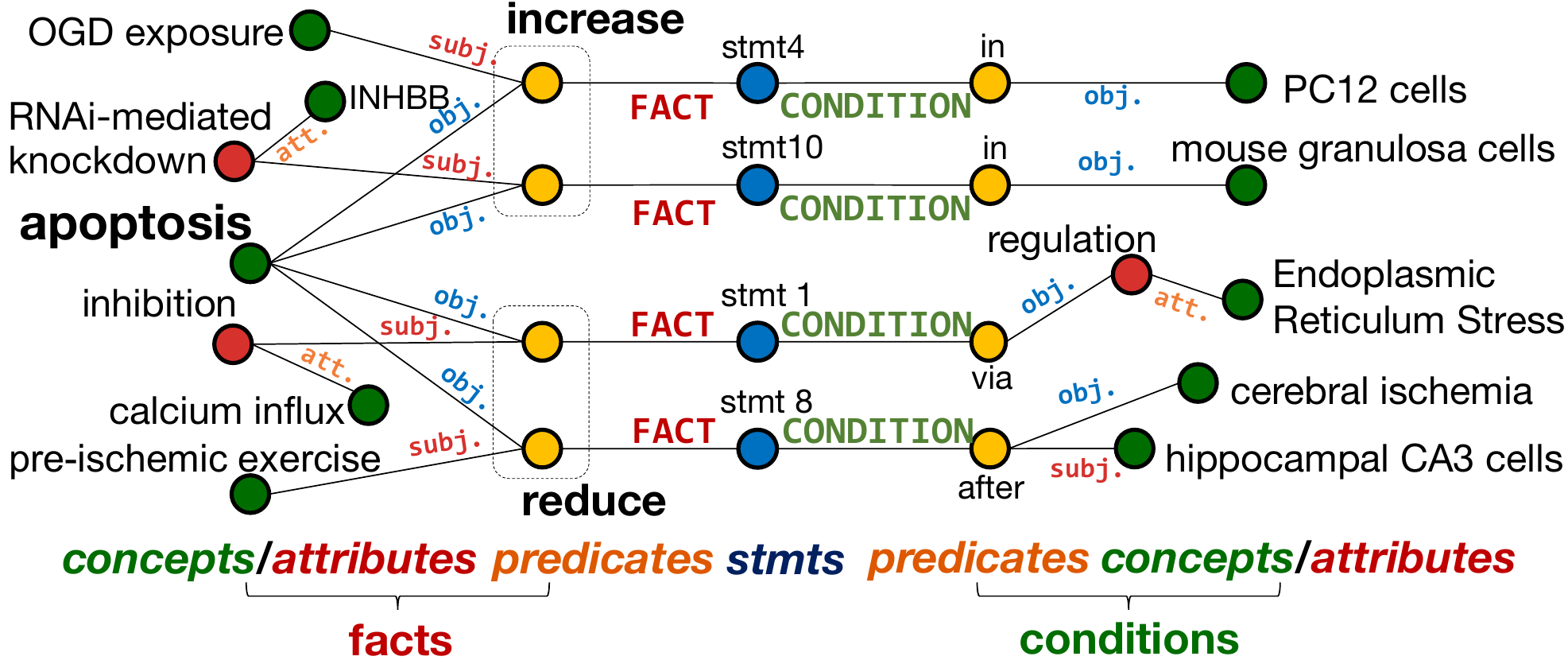}}
    \caption{Our biological knowledge graph provides a visualized, comprehensive understanding of what increased/reduced ``apoptosis'' under what kind of conditions.}
    \label{fig:case_study}
\end{figure}

We propose a new information extraction task named ``Biological Conditional Statement Structuring'' (BioCS). Given a biological statement sentence, BioCS extracts fact tuple(s) as well as condition tuple(s) on which the fact(s) were observed or claimed. The expected outputs of the example sentence are: \textbf{Fact 1:} (alkaline\_pH, increases, \{TRPV5/V6 \_channels: activity\}), and \textbf{Condition 1:} (TRPV5/V6\_channels, in, Jurkat \_T\_cells). The subject or object in the tuple is formatted as \{\underline{c}oncept: \underline{a}ttribute\} where the attribute could be null if it is a concept only. Actually we find another fact tuple from the full example sentence: \textbf{Fact 2:} (extracellular\_acidic\_pH, reduces, \{TRPV5/V6\_channels: activity\})




Inspired by \cite{stanovsky2018supervised}, we formulate BioCS as a sequence tagging problem. We propose a new deep sequence tagging framework to achieve the goal. Experiments on a data set of 141M sentences from PubMed paper abstracts show that the biological knowledge graph we constructed provide a good understanding of biological statements (\url{https://scikg.github.io}).

\section{Framework Description}
Our framework has two modules: (1) a deep sequence tagging model, taking multiple language feature sequences as inputs and returning multiple fact and condition tuples; (2) an iterative self-training scheme with massive unlabeled data to enhance the model.

\section{Preliminary Results}
Suppose we are interested in what \textit{increased}/\textit{decreased} ``apoptosis''. The KG provides us a snapshot in Figure~\ref{fig:case_study}: (1) ``OGD exposure'' and the ``RNAi-mediated knockdown'' of ``INHBB'' increased apoptosis, and (2) the ``inhibition'' of ``calcium influx'' and ``pre-ischemic exercise'' reduced apoptosis with the left side of the figure. It is important to be aware of the condition for each factual claim on the right side of the figure. They describe either the methodology of the observation or the context of it. This KG will enable effective biological knowledge inference and reasoning.


\bibliographystyle{ACM-Reference-Format}
\bibliography{main}


\begin{thebibliography}{3}


\ifx \showCODEN    \undefined \def \showCODEN     #1{\unskip}     \fi
\ifx \showDOI      \undefined \def \showDOI       #1{#1}\fi
\ifx \showISBNx    \undefined \def \showISBNx     #1{\unskip}     \fi
\ifx \showISBNxiii \undefined \def \showISBNxiii  #1{\unskip}     \fi
\ifx \showISSN     \undefined \def \showISSN      #1{\unskip}     \fi
\ifx \showLCCN     \undefined \def \showLCCN      #1{\unskip}     \fi
\ifx \shownote     \undefined \def \shownote      #1{#1}          \fi
\ifx \showarticletitle \undefined \def \showarticletitle #1{#1}   \fi
\ifx \showURL      \undefined \def \showURL       {\relax}        \fi
\providecommand\bibfield[2]{#2}
\providecommand\bibinfo[2]{#2}
\providecommand\natexlab[1]{#1}
\providecommand\showeprint[2][]{arXiv:#2}

\bibitem[\protect\citeauthoryear{Miller}{Miller}{1947}]%
        {miller1947nature}
\bibfield{author}{\bibinfo{person}{David~L Miller}.}
  \bibinfo{year}{1947}\natexlab{}.
\newblock \showarticletitle{The Nature of Scientific Statements}.
\newblock \bibinfo{journal}{\emph{Philosophy of Science}} \bibinfo{volume}{14},
  \bibinfo{number}{3} (\bibinfo{year}{1947}), \bibinfo{pages}{219--223}.
\newblock


\bibitem[\protect\citeauthoryear{Stanovsky, Michael, Zettlemoyer, and
  Dagan}{Stanovsky et~al\mbox{.}}{2018}]%
        {stanovsky2018supervised}
\bibfield{author}{\bibinfo{person}{Gabriel Stanovsky}, \bibinfo{person}{Julian
  Michael}, \bibinfo{person}{Luke Zettlemoyer}, {and} \bibinfo{person}{Ido
  Dagan}.} \bibinfo{year}{2018}\natexlab{}.
\newblock \showarticletitle{Supervised open information extraction}. In
  \bibinfo{booktitle}{\emph{NAACL}}. \bibinfo{pages}{885--895}.
\newblock


\bibitem[\protect\citeauthoryear{Tomilin, Cherezova, Negulyaev, and
  Semenova}{Tomilin et~al\mbox{.}}{2016}]%
        {tomilin2016trpv5}
\bibfield{author}{\bibinfo{person}{Victor~N Tomilin}, \bibinfo{person}{Alena~L
  Cherezova}, \bibinfo{person}{Yuri~A Negulyaev}, {and}
  \bibinfo{person}{Svetlana~B Semenova}.} \bibinfo{year}{2016}\natexlab{}.
\newblock \showarticletitle{TRPV5/V6 channels mediate Ca2+ influx in jurkat T
  cells under the control of extracellular pH}.
\newblock \bibinfo{journal}{\emph{Journal of cellular biochemistry}}
  \bibinfo{volume}{117}, \bibinfo{number}{1} (\bibinfo{year}{2016}),
  \bibinfo{pages}{197--206}.
\newblock


\end{thebibliography}


\end{document}